\documentclass[journal]{IEEEtran}

\usepackage{cite}
\usepackage{url}
\ifCLASSINFOpdf
  \usepackage[pdftex]{graphicx}
  \graphicspath{{./figure/}}
  \usepackage[caption=false,font=footnotesize]{subfig}
\else

\fi

\usepackage{mathtools}
\usepackage{amssymb}
\usepackage{bm}
\usepackage[ruled,vlined,linesnumbered]{algorithm2e}

\usepackage{url}

\usepackage{microtype}

\usepackage{todonotes}
\usepackage{color}
\usepackage{caption}  

\begin{document}

\title{Artificial intelligence empowered multi-AGVs in manufacturing systems}

\author{Dong Li, Bo Ouyang, Duanpo Wu, Yaonan Wang
\thanks{This work was supported by the National Natural Science Foundation of China under Grant No. 61603131, and the Natural Science Foundation of Hunan Province under Grant No. 2017JJ3036.}%
\thanks{Dong Li, Bo Ouyang, and Yaonan Wang are with the College of Electrical and Information Engineering, Hunan University, Changsha 410082, China (e-mail: \{201313010310, ouyangbo, yaonan\}@hnu.edu.cn).
Dong Li is also with the College of Biomedical Engineering \& Instrument Science, Zhejiang University, Hangzhou 310027, China.}%
\thanks{Duanpo Wu is with the School of Communication Engineering, Hangzhou Dianzi University, Hangzhou 310018, China (e-mail: wuduanpo@hdu.edu.cn). He is also with Hangzhou Neuro Science and Technology Co., Ltd., Hangzhou 310052, China.}%
}	


\maketitle

\begin{abstract}
AGVs are driverless robotic vehicles that picks up and delivers materials.
How to improve the efficiency while preventing deadlocks is the core issue in designing AGV systems.
In this paper, we propose an approach to tackle this problem.
The proposed approach includes a traditional AGV scheduling algorithm, which aims at solving deadlock problems, and an artificial neural network based component, which predict future tasks of the AGV system, and make decisions on whether to send an AGV to the predicted starting location of the upcoming task, so as to save the time of waiting for an AGV to go to there first when the upcoming task is created.
Simulation results show that the proposed method significantly improves the efficiency as against traditional method, up to 20\% to 30\%.
\end{abstract}


\IEEEpeerreviewmaketitle

\section{Introduction}
\label{sec:intro}

\IEEEPARstart{R}{ecent} years have seen increasing popularity of AGVs (Automated Guided Vehicles) in industrial applications.
AGVs are robotic vehicles that picks up and delivers materials around a manufacturing facility or warehouse.
The introduction of AGVs has brought many notable advantages\cite{sabattini2013technological}, such as the production efficiency and flexibility.

At the heart of AGV system design is the intention to improve the efficiency while preventing deadlocks.
In \cite{reveliotis2000conflict}, Reveliotis proposed a multi-AGV control architecture where the deadlock resolution and the system performance considerations are decoupled.
An efficient algorithm was also proposed to constrain the whole system in safe states, so as to rule out deadlock problems.
Petri nets are another popular mathematic tool to handle deadlock problems because of their inherent characteristics, leading to a variety of deadlock-control policies in AGV systems \cite{wu2004modeling,wu2007shortest}.

While the deadlock problem receives a lot of attention in the past few decades, much less literatures focus on how to improve the efficiency.
\cite{Confessore2013A} attempt to optimize the dispatch rules by relying on the formulation of a Minimum Cost Flow Problem to improve the efficiency. Similarily, in \cite{Singh2011AGV,Lin2006Network}, the authors optimize the dispatch rules to improve the efficiency by monitoring parameters reﬂecting efﬁciency and uniformity of material distribution.
In \cite{digani2015quadratic}, the entire factory operation environment is divided into several sectors, so as to reduce the number of traffic conditions such as AGV encounters, and eventually improve the efficiency.
Shortest paths are not necessarily the the least time-consuming routes for AGVs to take. 
In \cite{sa2015mission,Zhang2016Dynamic,Chen2017Research}, the current traffic is taken into account in path planning, so as to schedule paths with shortest travel time to AGVs.
Many heuristic methods are proposed to directly compromise deadlock preventing requirements and the desire for efficiency \cite{gen2014multiobjective}.

Usually, in industrial environments, the AGVs work together with humans and many other machines.
But these machines are often designed and installed independently.
Unaware of other systems is a source of inefficiency.
In this paper, we propose a novel method to improve the efficiency of AGVs by predicting future tasks.
Usually, tasks are created by operators, demanding an AGV to pick up materials from one location and then deliver to another.
If the central coordinator knows exactly where the next task begins, it can make use of this information and send an idle AGV to that location, so that the next task can be executed immediately once it is created.
The reason that it is possible for the central controller to make the prediction is that usually tasks are not arbitrary in most applications, i.e., tasks are usually correlated.
For example, in assembly lines the the semi-finished products are moved from one workstation to the next.
Correspondingly, if an AGV system is adopted, tasks demanding AGVs to deliver products to consecutive workstations are likely to be highly correlated.

Deep learning, a rapidly growing field in the past decade, has shown to be very effective in variety of prediction tasks.
One category of deep learning models, namely the recurrent neural networks with Long Short-Term Memory (LSTM) \cite{hochreiter1997long},\cite{sutskever2014sequence} are exceptionally outstanding for handling sequences \cite{Gers2000Learning,gers2000recurrent,gers2001lstm}, and is utilized in this paper.
Deep learning is sometimes criticized for being a ``black box'' \cite{marcus2018deep}, especially in applications where the performance should be insured, e.g. industrial, medical applications, etc.
For this reason, in this paper, we are not going to propose an approach to coordinate AGVs by deep learning alone, but rather, we combine traditional AGVs scheduling algorithms with deep recurrent neural networks, and take advantage of both methods.
With traditional scheduling algorithms, for example, dynamic path scheduling algorithm based on time window (hereinafter referred to as DPSTW)\cite{smolic2009time} or greedy path scheduling algorithm (hereinafter referred to as GREEDY, which is adopted in \cite{opentcs}, an open source platform), deadlock and conflicts are strictly prevented, while with deep learning, the efficiency is increased as we can predict future tasks.

\begin{figure*}
\label{fig:perf}
\subfloat[Guidepath 1]{\includegraphics[width=0.24\textwidth]{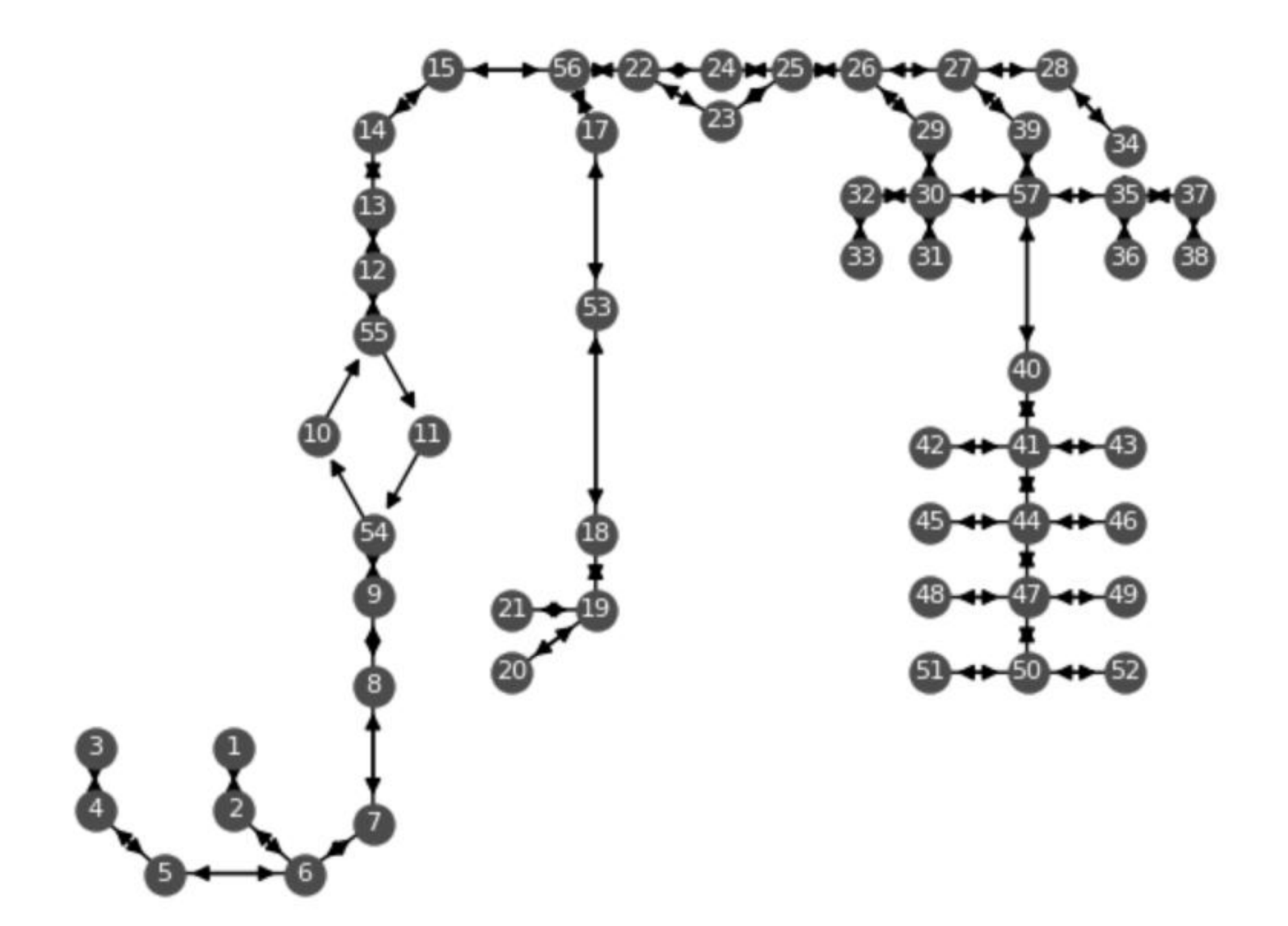}}
\subfloat[Guidepath 2]{\includegraphics[width=0.24\textwidth]{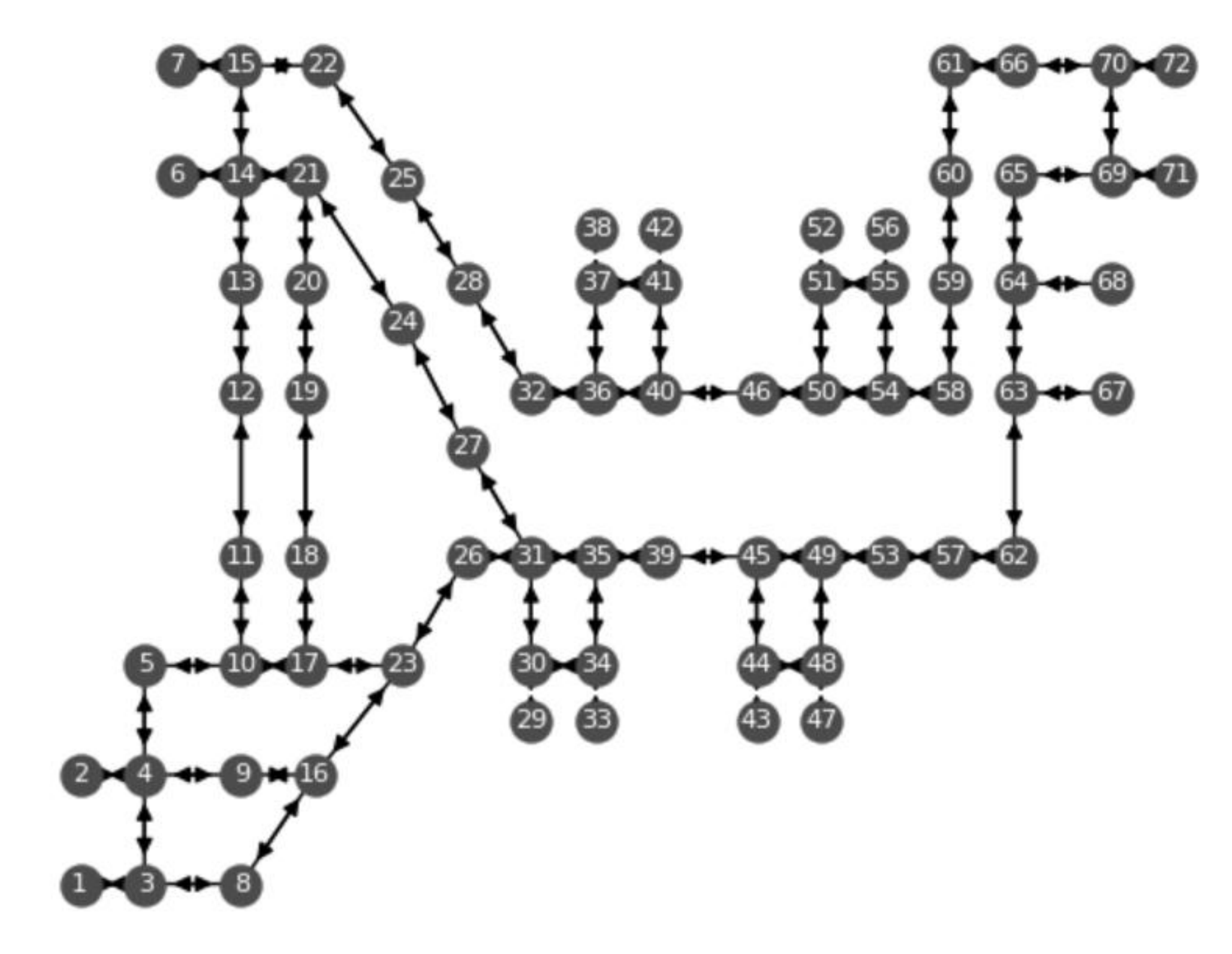}}
\subfloat[Guidepath 3]{\includegraphics[width=0.24\textwidth]{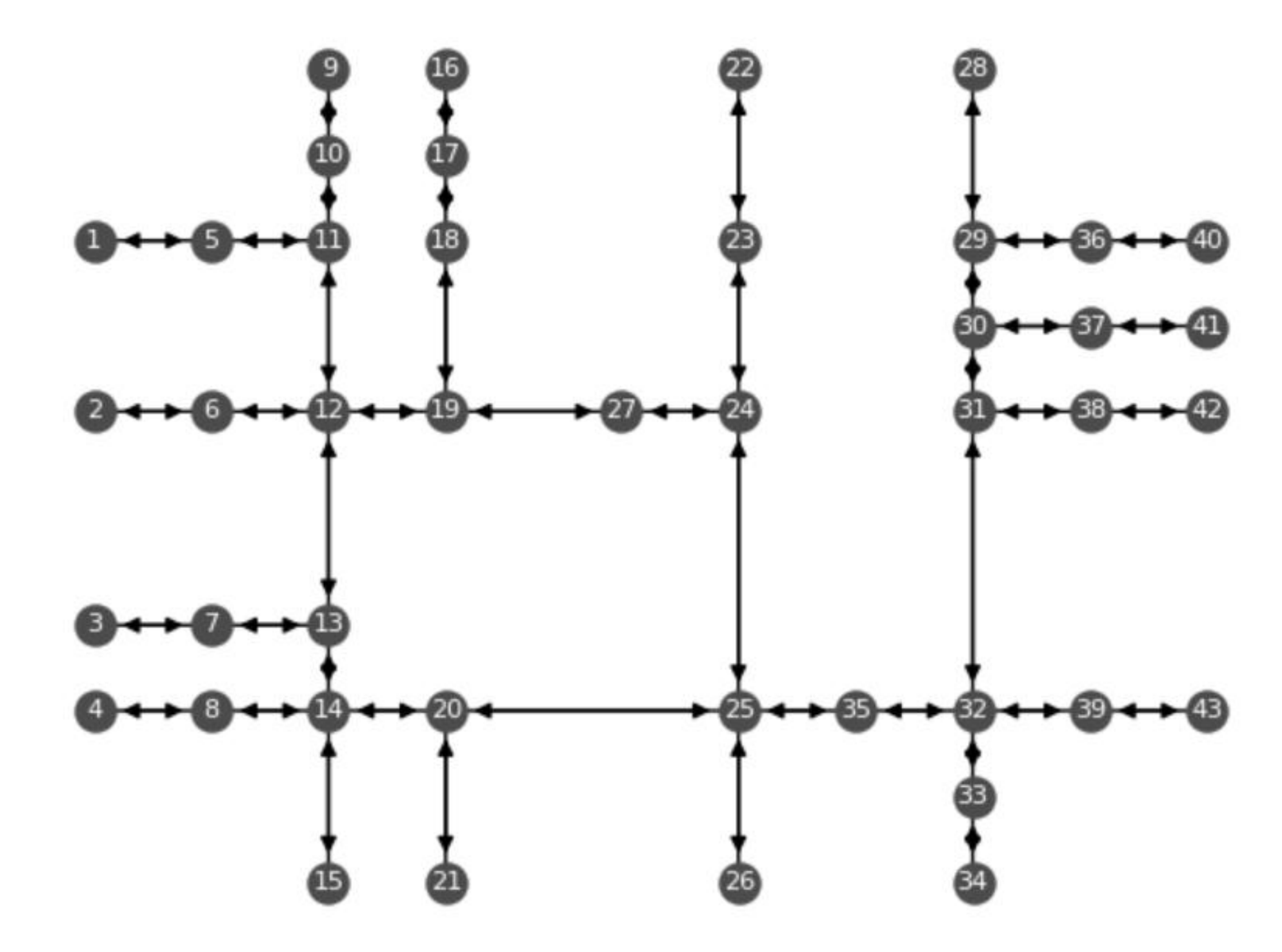}}
\subfloat[Guidepath 4]{\includegraphics[width=0.24\textwidth]{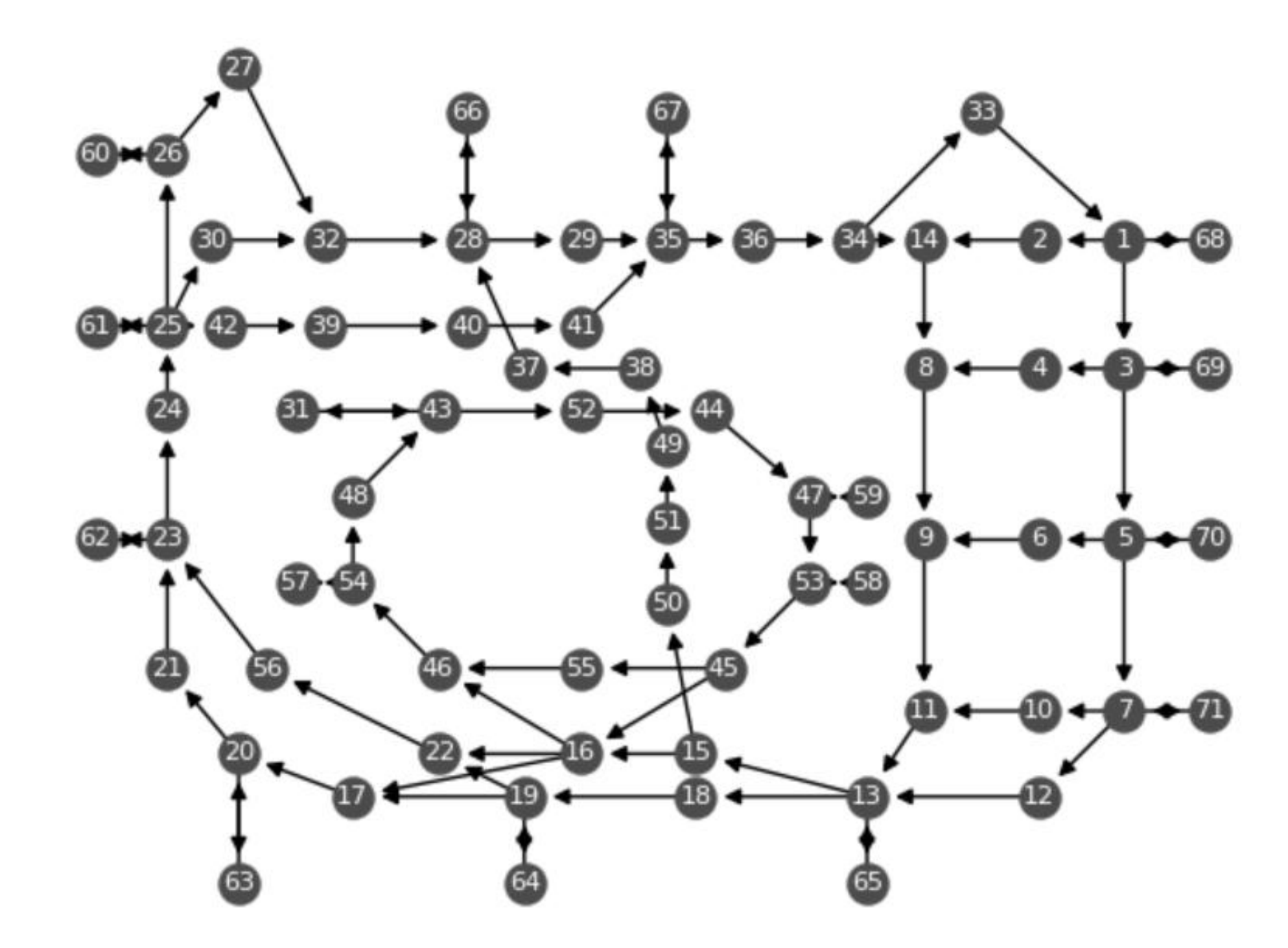}}
\caption{Guidepath graphs.}
\label{fig:graph}
\end{figure*}

The rest of the paper is organized as follows. 
In the Section \ref{sec:notations}, several notations are introduced. 
Section \ref{sec:coord} describes the DPSTW and GREEDY. 
How we use neural network to predict the future task is introduced in Section \ref{sec:predict}.
In Section \ref{sec:efficiency} we propose a method to make use of the prediction result to increase the efficiency of the AGV system.
Simulation results are then shown in Section \ref{sec:experiment}. 
Finally, we contain some concluding remarks in Section \ref{sec:conclusion}.
 
\section{Notations}
\label{sec:notations}

The guidepath of a considered AGV system is defined as a directed graph $\mathrm{G}=\{\mathrm{V}, \mathrm{E}\}$ (shown in Fig.1) which contains a set of nodes $\mathrm{V}$ and a set of weighted edges $\mathrm{E}$. 
A node represents the intersection or the end of a path, while an edge represents a section of a path between adjacent nodes.
The weight of an edge denotes its length or the nominal travel time of AGVs on it.

Assume that there are $N$ AGVs travel on $\mathrm{G}$.
Each of them are either idle or processing tasks.
Tasks are created by operators, demanding an AGV to travel from a starting node to a destination node, and possibly perform some actions there.

The set of all tasks is denoted by $T$.
All the completed tasks consists of a set $C$.
The active tasks are then within the set $\mathrm{Q}=T\setminus C$.



\section{AGV Coordination}
\label{sec:coord}
Functionally, the central coordinator consists of three major components, which are the router, the dispatcher, and the scheduler.

The main purpose of the router is to find one or more routes for a given pair of nodes.
When an AGV is assigned a task, it needs to find a route first.
If the router only finds the shortest route, it might take a long time for all the tasks to complete since some of the routes may overlap and at least one AGV has to wait until the overlapped part of routes is released.
However, if several alternative routes are given, it is possible to find a more efficient scheduling result. 
The router adopts Yen's algorithm \cite{yen1971finding} to accomplish this goal, which tries to find $k$ shortest routes for the given pair of nodes.
Yen's algorithm is based on Dijkstra's algorithm, the latter of which finds only one shortest path and can be viewed as a special case ($k=1$) of the former.

\begin{figure}[t]
\centering
\includegraphics[width=0.4\textwidth]{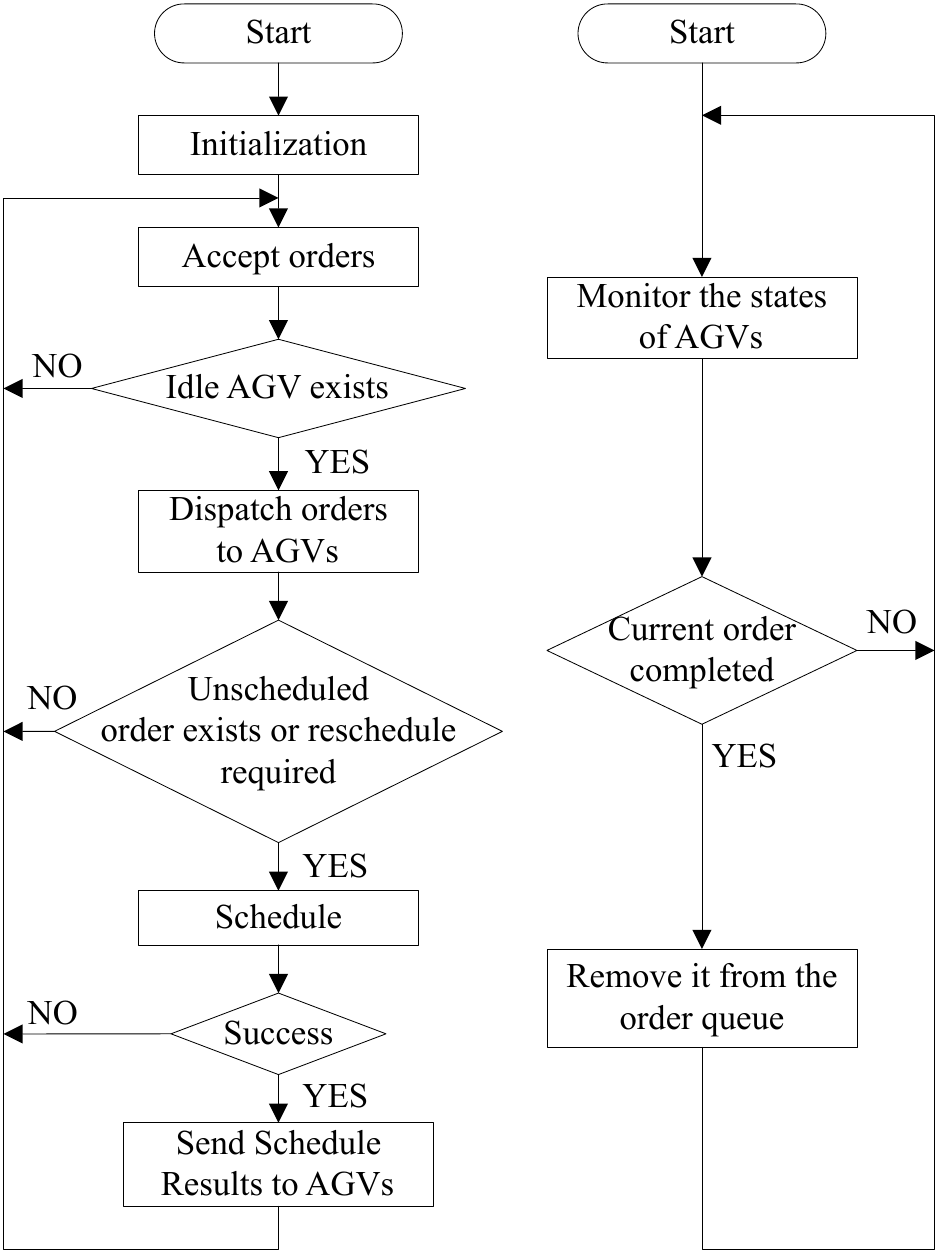}
\caption{Dispatcher and scheduler implementation. Left: Task dispatching and scheduling thread. Right: Monitoring thread.}
\label{fig:flow_chart}
\end{figure}

The dispatcher decides which order should be processed by which AGV. 
This is accomplished by a greedy algorithm.
The tasks are first sorted according to their priorities.
Then, the dispatcher attempts to assign the first unassigned task to an idle AGV that is nearest to the starting node of the task.
The distance between the current location of the AGV and the starting node is given by the router.
If all AGVs are busy, the tasks just hang on and wait for the dispatched orders to be completed.

The scheduler allocates resources for the dispatched orders.
Two different algorithms are considered in this article.

The first one is a time-window based algorithm \cite{smolic2009time} (Dynamic Path Scheduling algorithm based on Time Window, referred to as DPSTW in the following), which prevents conflict and deadlock problems in advance.
Specifically, the scheduler determines at what time, which robot will occupy which arc.
Thus it is able to prevent the deadlock and conflicts at scheduling time.
The robots are first sorted according to the priorities of the orders they are processing.
Then the scheduler tries to register a feasible earliest time-window for each arc these robots are going to travel, one after another.
The width of the time-window is determined by the length of the arc and the velocity of the robot.
Let us elaborate the scheduling process by an example.
Assume that some arc is occupied at time intervals $(t_1, t_2)$ and $(t_3, t_4)$.
At time $t_0$, we are going to register a time-window on this arc with width $w$.
If $t_1-t_0>w$, it is inserted before $(t_1, t_2)$.
If this is not the case, then the length of the intervals between these two intervals is checked.
If the condition $t_3-t_2>w$ is satisfied, the time-window is inserted between $(t_1, t_2)$ and $(t_3, t_4)$.
Otherwise, it is inserted after $(t_3, t_4)$.
For more details, we refer the reader to \cite{smolic2009time}.

Another scheduling algorithm is a greedy algorithm.
In this algorithm, if a robot is about to enter an arc, it will first asks the edge controller to check if the arc and its ending nodes are being occupied by other robots.
If not, entering the edge is approved.
Otherwise, the robot has to wait, until the resources are freed.
As one can easily pointed out, deadlock prevention is not guaranteed in this algorithm. 
But under certain circumstances, such problems can be avoided.
A typical example is that the graph $G$ is circular, with no arcs pointing in opposite directions.
In such a graph, all robots travel in the same direction and therefore there will be no deadlock problems.
This simple algorithm is actually effective, if it is possible to design such a graph.
It is adopted in OpenTCS \cite{opentcs}, an open source AGV control system.


\section{PREDICTING FUTURE TASKS}
\label{sec:predict}

As we have mentioned, deep learning methods are very efficient in tasks such as prediction.
In this section, we develop a deep learning model to efficiently predicting future tasks.
To begin with, assume that we have a data set of recorded tasks, each of which were created by the operators.
Consecutive tasks, or, a sequence of tasks in the data set, are utilized to make a prediction of the next task.
To be more specific, the sequence of starting nodes of the past $R$ tasks are utilized to predict the staring node of the next task.

\begin{figure}[t]
\centering
\includegraphics[width=3.4in]{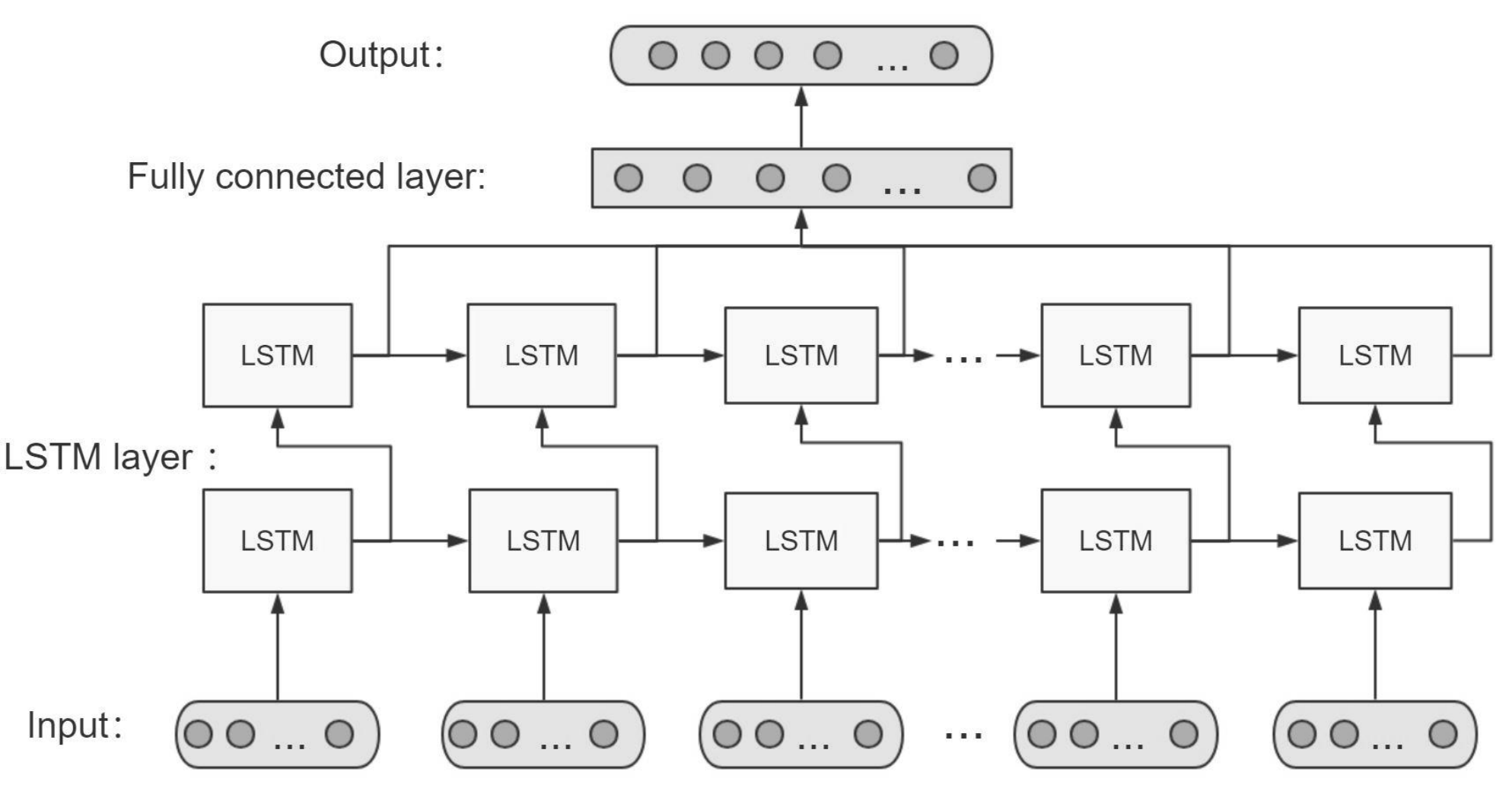}
\caption{The structure of the neural network}
\label{fig:rnn}
\end{figure}

As sequential data is involved, it is natural to use Recurrent Neural Networks (RNNs) to build the model.
Our neural network model is shown in Fig.~\ref{fig:rnn}.
The input $[\delta]$ consists of the starting nodes of the past $R$ tasks, as we have stated.
Each node is represented by an one-hot vector, that is, a vector filled with 0 except for a 1 at the index of the node.
The input is connected to a two layer LSTM cells, whose output are then fed into a fully connected layer.
The topmost is an output layer, computing the logit which predict the log-likelihood of the starting node of the next task, as shown in Fig.~\ref{fig:rnn}.

\section{INCREASING EFFICIENCY}
\label{sec:efficiency}
Thanks to the model introduced in the last section, we are able to predict the starting node $s$ of the next task $t$.
In this section we propose a method to make use of this information to increase the efficiency of the system.

The approach we are going to take is creating a new task $p$ with $s$ being its target node, that is, by creating the new task $p$, we demand the system to send an AGV to $s$.
The intuition is that once the next task $t$ is created by operators, it can be immediately executed by the AGV we sent to $s$, without having to wait for some AGV to go there first.
For convenience, let us call task $p$ the predicted task.
In other words, predicted tasks are tasks created by the algorithms that are described in this section, so that they can be distinguished from tasks directly created by operators.

The first problem we need to solve is determining the best timing for creating predicted tasks.
Continuously creating predicted tasks will increase the system overhead, and is more likely to reduce the efficiency.
A natural method to solve this problem is creating predicted tasks only when the system is not busy.
To measure how idle the system is,we define a measure
\begin{equation}
\label{eq:idle}
\begin{split}
idle = & \frac{\text{average duration between created tasks}}{\text{average completing time of tasks}} \\
= & \frac{\sum_{i\in C} [\tau_{end}(i)-\tau_{start}(i)]/|C|}{\tau /|T|} \;.
\end{split}
\end{equation}
$\tau$ is the elapsed time of the system, and recall that $C$ is the set of all completed tasks, and $T$ contains all the tasks that have been created.
If $idle = 1$, then on average, a task is completed during the time a new task is created.

Note that we do not simply measure how idle the system is by counting the number of idle AGVs, because it is possible that the dispatcher and the scheduler are buffering tasks, and they will be assigned to these AGVs instantly. 
We can combine both the number of idle AGVs $n$, and the $idle$ measure to determine a good timing for creating predicted tasks.
If at present, the following condition is satisfied
\begin{equation}
\label{eq:cond}
\left\{\begin{array}{ll}{n \geq n_1} & {\text{if } idle < 0.8} \\ {n \geq n_2} & {\text{if } 0.8\le idle < 1.2}\\ {n \geq n_3} & {\text{if } 1.2\le idle < 1.6}\\ {n \geq n_4} & {\text{if } idle > 1.6}\end{array}\right.
\end{equation}
then a predicted task is created, where $n_1, n_2, n_3$ and $n_4$ are hyperparameters.
They can be tuned by cross-validation.

\begin{algorithm}
\caption{optimization operation} 
\label{alg2}

\If{a new task $t$ is created}
{
    Append starting node of $t$ to the end of sequence $seq$\;
    \If{$seq$'s length is greater than $R$}
    {
    	  Remove the first element from $seq$.
    }
    
    \If{prediction is wrong}
    {
         Append  $t$ to $Q$\;
         \If{$p$ is being processed}
         {
            Set the status of the vehicle $v$ which is processing $p$ to idle\;
         }
         Delete $p$ from $Q$ \;
    }
    \Else
    {
      \If{$p$ \textbf{is not} completed}
      {
         Append $t$ to vehicle $v$'s task queue\;
      }
    }
     \If {$seq$'s length equals $R$ and condition \eqref{eq:cond} is satisfied}
     {
        $s \gets PredictStartingNodeOfNextTask([\delta])$\;
        Create a new task $q$ with $s$ being its destination node\;
        Append $q$ to $Q$\;
     }
}
\end{algorithm}

\begin{figure*}
\subfloat[Guidepath 1]{\includegraphics[width=0.48\textwidth]{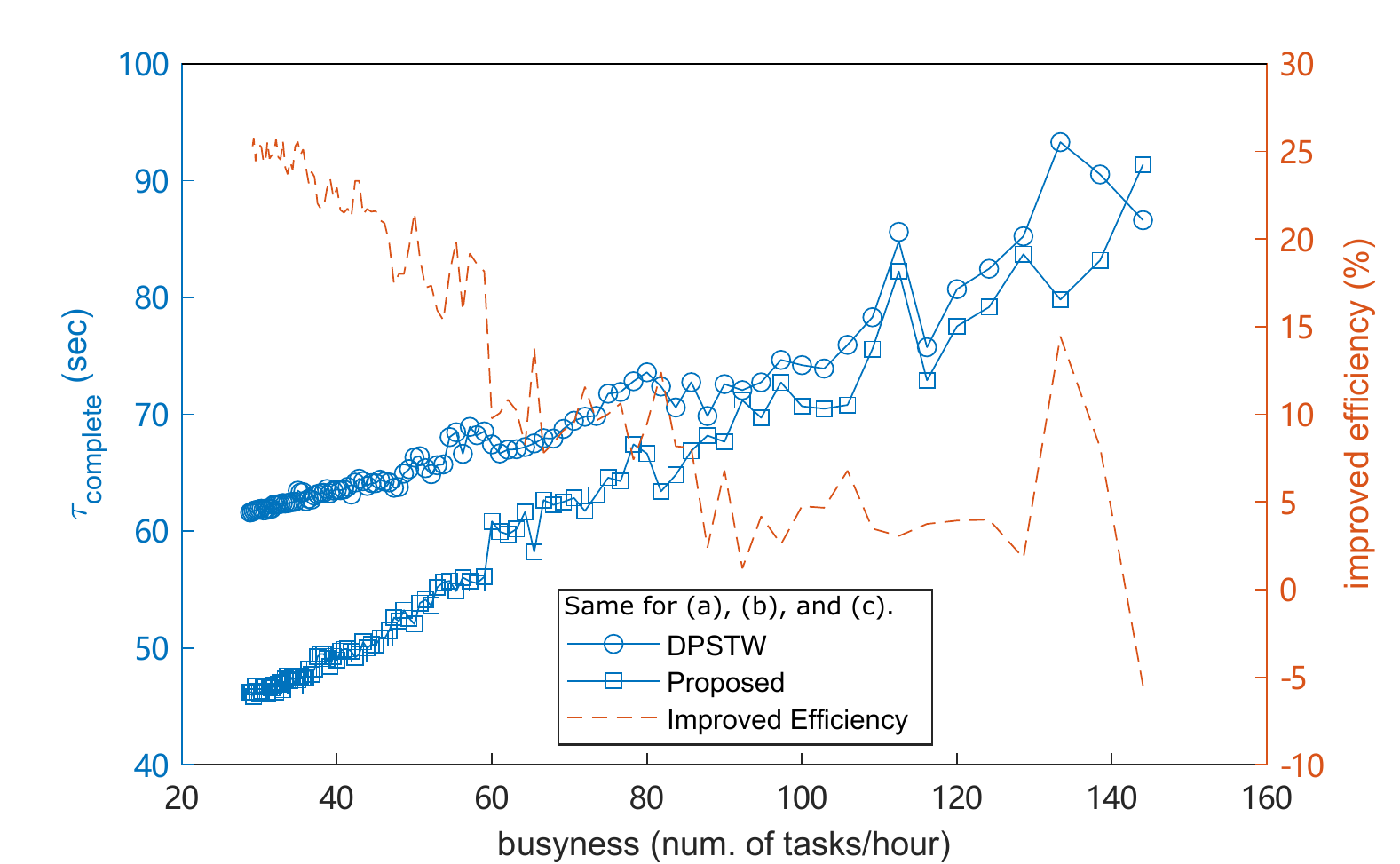}}
\subfloat[Guidepath 2]{\includegraphics[width=0.48\textwidth]{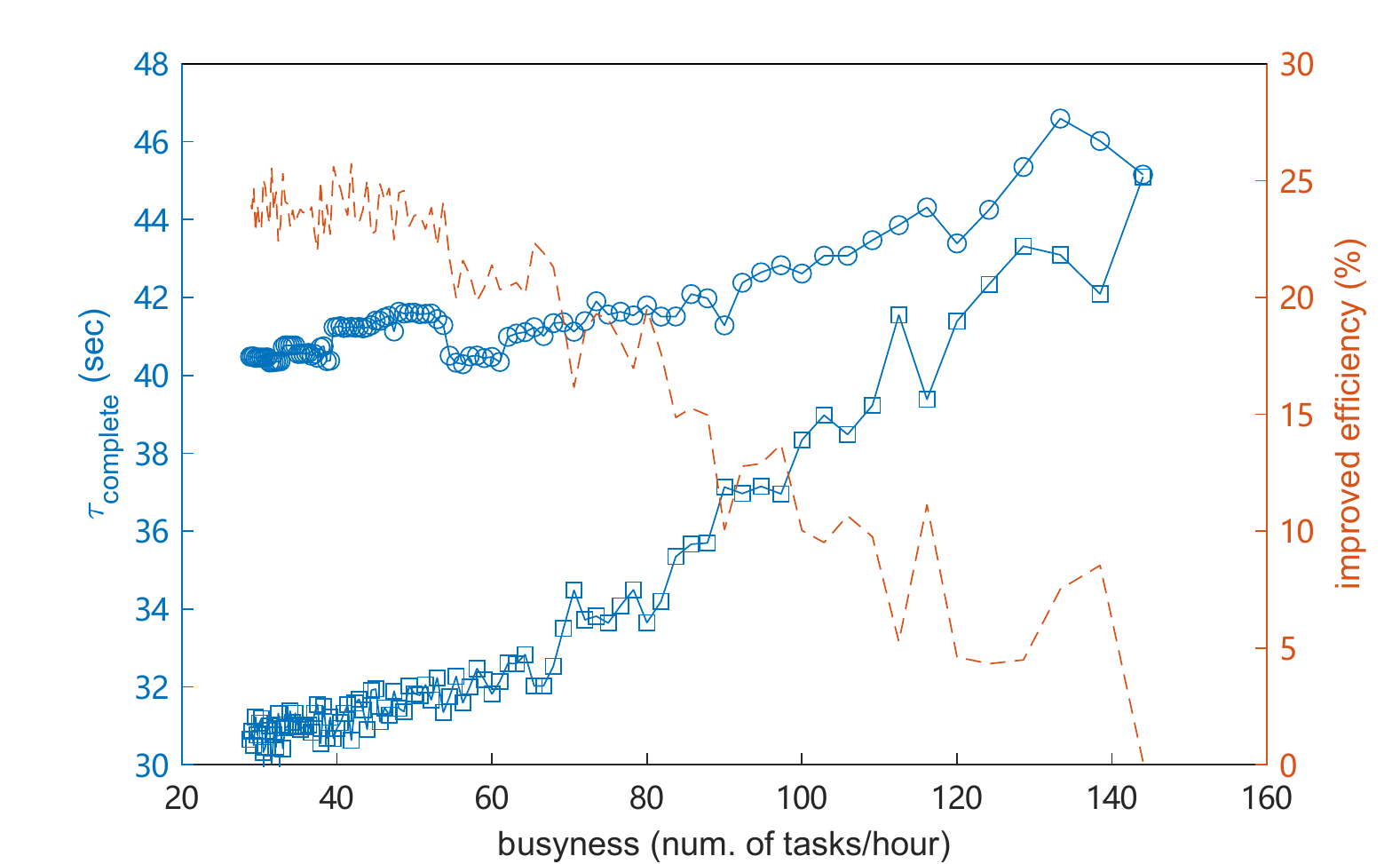}}
\\
\subfloat[Guidepath 3]{\includegraphics[width=0.48\textwidth]{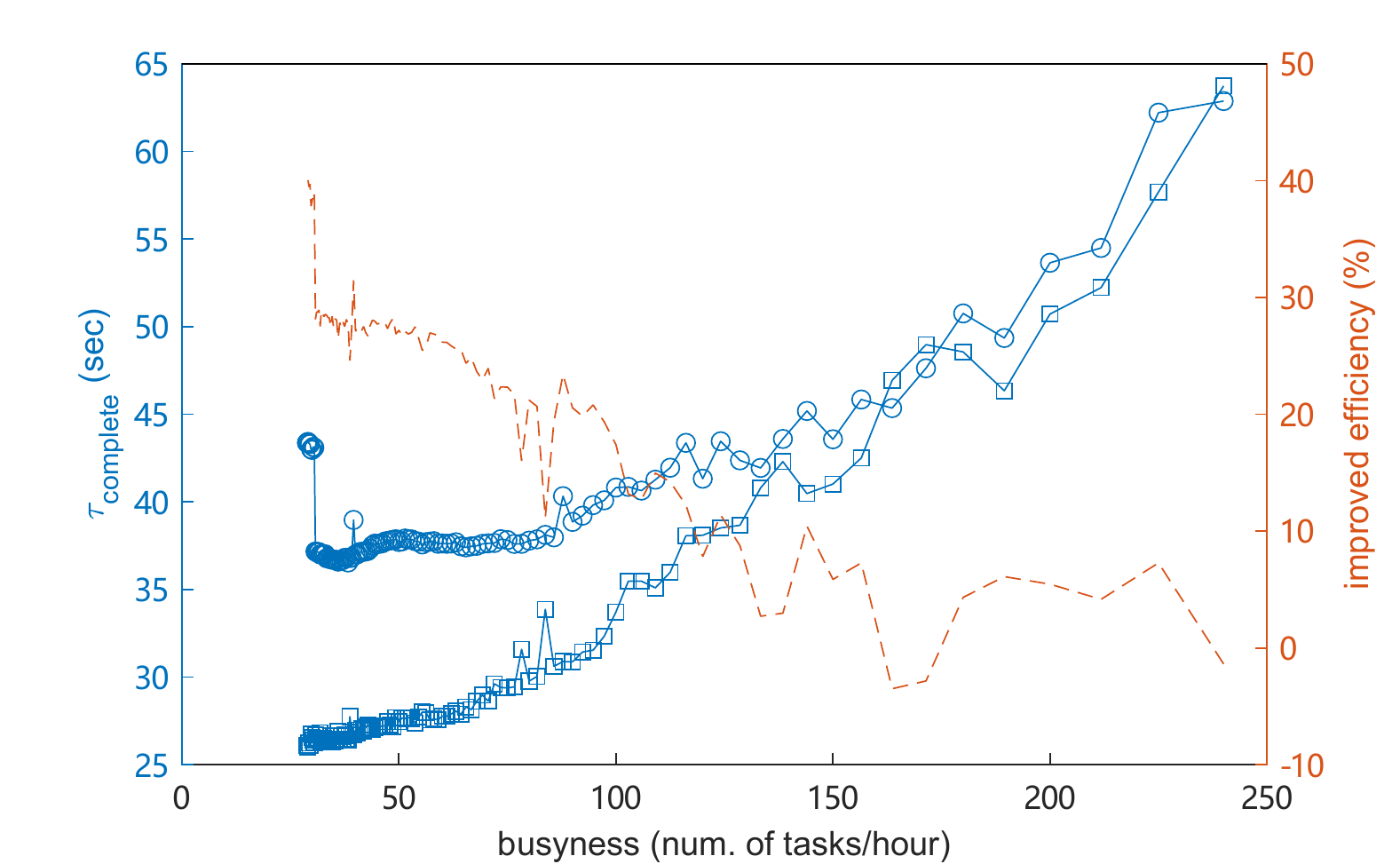}}
\subfloat[Guidepath 4]{\includegraphics[width=0.48\textwidth]{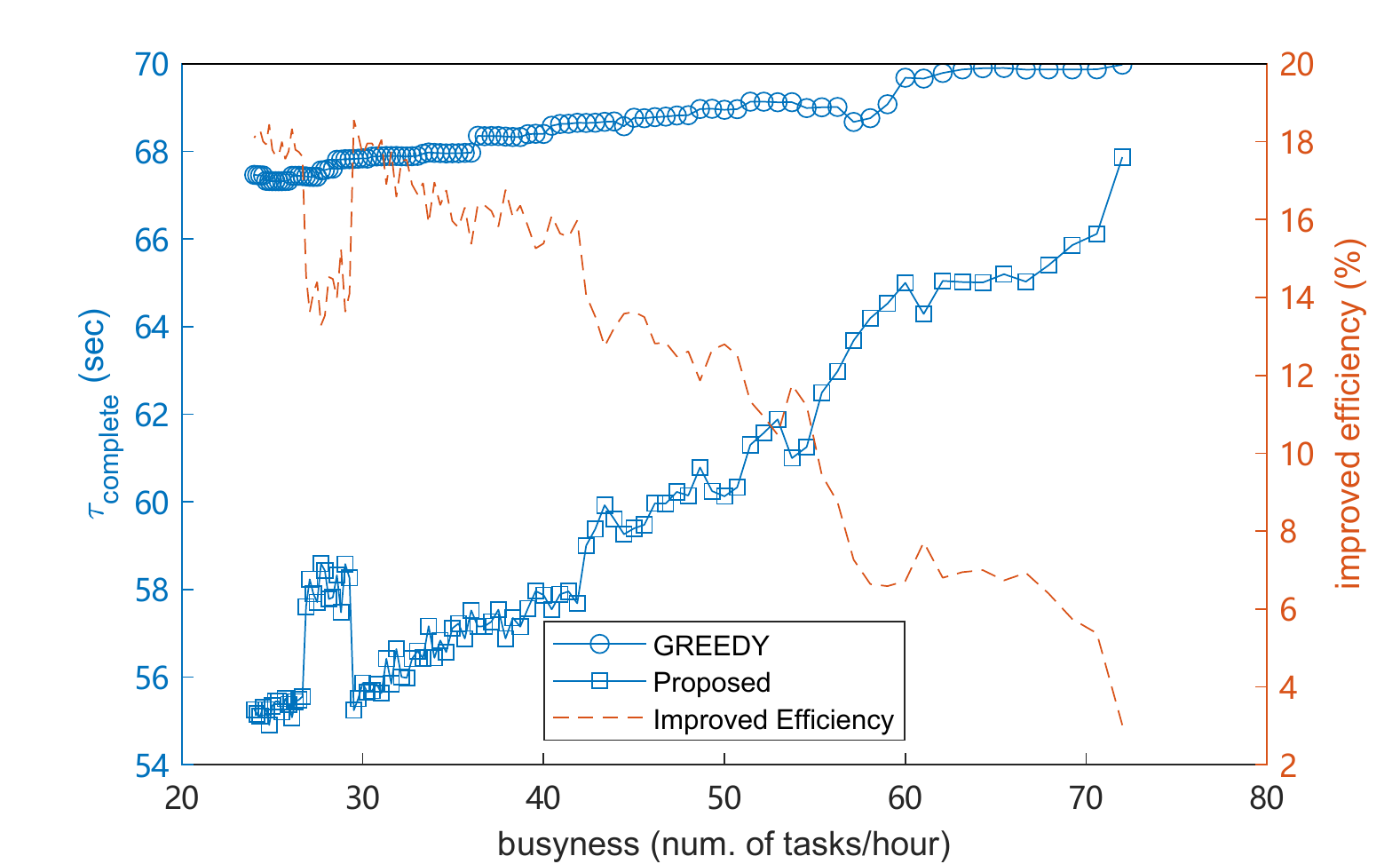}}
\caption{Performance evaluation.}
\label{fig:result}
\end{figure*}

Now another problem is that the predicted starting node $s$ of the next task might be wrong.
In such cases, sending an AGV to $s$ only increases the overhead of the system.
To overcome this problem, if a predicted task $p$ is created, we first determine whether the prediction was wrong or not by comparing it with the next task $t$ from the operator.
Note that this can not be done until $t$ is created.
If it was wrong, we just cancel the predicted task $p$ immediately.
If we made a correct prediction, the newly created task $t$ is manually assigned to the AGV processing $p$.
So that the AGV will excecute $t$ right after it finishes task $p$.

\section{Simulation Results}
\label{sec:experiment}

To evaluate the performance of the proposed method, we conduct several experiments with different guidepaths, i.e., each of the experiments has its own configurations of markers and wires. 
The guidepaths considered in the simulation are shown in Fig.~\ref{fig:graph}.
Two of the guidepaths are actually being used in real plants (Guidepath 1 and 2), and the other two are synthesized, which look similar to a grid and a ring, respectively.
We combine the proposed method with two conventional scheduling algorithms, namely, the DPSTW and the GREEDY algorithm, to show how the efficiency is improved.
Since in the GREEDY algorithm deadlock problems may happen, the ring-like guidepath (Guidepath 4) is designed specifically for the purpose that no robots will travel in opposite directions.
So that the deadlock problem is prevented.
The number of AGVs is set to 8.

In real cases, tasks are correlated.
The correlation is modeled by a Markov chain.
The transition matrix is defined as
\begin{equation}
\label{eq:transition}
P_{i,j}=\Pr (s_{k+1}=i|s_k=j) \;,
\end{equation}
where $s_k$ and $s_{k+1}$ are the starting nodes of consecutive tasks.
In the simulation, the tasks are generated according to Eq.~\eqref{eq:transition}, i.e., their starting nodes are correlated as indicated by Eq.~\eqref{eq:transition}.
The busyness of the system is defined to be the average number of tasks issued in an hour. 
These generated tasks are then divided into training set $D_{train}$ and test set $D_{test}$, and then used to train the neural network and evaluate the performance respectively.
The average completion time $\tau _ {complete}$ of all tasks is used to evaluate the performance:
\begin{equation}
\tau _ {complete}=\frac{1}{|D_{test}|}  \sum_{t\in D_{test}}\left(\tau_{end}(t)-\tau_{start}(t)\right) \;.
\end{equation}

The average completion time of tasks for the first three guidepaths are shown in Fig.~\ref{fig:result} a, Fig.~\ref{fig:result} b and Fig.~\ref{fig:result} c. 
In these figures, the curves marked with circles and squares are the average completion time for DPSTW and the proposed method, respectively. The dashed curve shows how the efficiency is improved in the proposed algorithm, with respect to DPSTW.
In the Fig.~\ref{fig:result} d, we compare the proposed method with the GREEDY algorithm, accordingly the map we use is the last guidepath(shown in Fig.~\ref{fig:graph} d.

We can see that the proposed method is more efficient than DPSTW and GREEDY in most cases.
Only when the system becomes really busy, their performance get close.
For example, in Fig.~\ref{fig:result} c, when almost 250 tasks are issued in an hour, or, new tasks are issued every 14 seconds, the improved efficiency approaches 0.
This makes sense because in such cases not too much can be done to improve the efficiency.
As the system becomes busy, the performance of DPSTW, GREEDY and the proposed method all decays.
But in cases of practical significance, with the proposed method, the efficiency is improved, by up to 20\% to 30\%.



\section{Conclusion}
\label{sec:conclusion}
This paper presents an approach to predict future tasks in multi-AGV systems.
This approach takes advantage of both traditional scheduling algorithms and deep learning.
With the former, deadlock and conflicts are strictly prevented, while with deep learning, the efficiency is increased.
By predicting the starting node of the upcoming task and scheduling an AGV there if certain conditions (Eq.~\eqref{eq:cond}) are satisfied, it saves the time of waiting for an AGV to go to the starting node when the upcoming task is created. 
The efficiency is shown to be significantly improved.
\ifCLASSOPTIONcaptionsoff
  \newpage
\fi
\bibliographystyle{IEEEtran}

\bibliography{ref}

\end{document}